%% file: example.tex
\documentclass{article}

\usepackage[final]{corl_2026} 
\usepackage{amsmath}
\usepackage{graphicx}
\usepackage{multirow} 
\usepackage{wrapfig}
\usepackage{subcaption}
\usepackage{xcolor}
\usepackage{algorithm}      
\usepackage{algpseudocode}  

\title{EagleVLA: Towards Onboard Real-Time Robot Control via Foresight-Aligned Asynchronous Inference}

\author{
\bfseries Zebin Yang$^{1,2}$ \quad Qi Wang$^{1}$ \quad Yunhe Wang$^{1}$ \quad Xiurui Guo$^{4}$ \quad Bo Yu$^{3}$ \\[0.4em]
\bfseries Shaoshan Liu$^{3}$ \quad Jiafeng Xu$^{1}$ \quad Hao Dong$^{1}$ \quad Meng Li$^{1,2*}$
\\[0.4em]
$^{1}$Peking University
~
$^{2}$TLAIC
~
$^{3}$AIRS
~
$^{4}$PrimeBot Research Institute
\\[0.2em]
$^{*}$ Corresponding author. Email: meng.li@pku.edu.cn
}

\begin{document}
\maketitle


\input{corl_2026_template_submission/Docs/1abstract}
\input{corl_2026_template_submission/Docs/2intro}
\input{corl_2026_template_submission/Docs/3background}

\input{corl_2026_template_submission/Docs/4challenge}

\input{corl_2026_template_submission/Docs/5method}

\input{corl_2026_template_submission/Docs/6experiments}

\input{corl_2026_template_submission/Docs/7conclusion}

\bibliography{example}  
\newpage
\appendix
\input{corl_2026_template_submission/Docs/8appendix}

\end{document}

%% file: corl_2026_template_submission/Docs/1abstract.tex
\begin{abstract}
    Vision-Language-Action (VLA) models have achieved impressive performance on diverse embodied tasks. However, deploying VLA models on low-power onboard devices, such as the Jetson Orin, remains challenging due to their high computational complexity, which leads to substantial inference latency and low control frequency. Asynchronous inference can partially mask this latency by parallelizing action execution and subsequent inference, but it introduces two critical issues: perception-execution misalignment and long reaction time. In this paper, we propose EagleVLA, a method for efficient VLA deployment on onboard devices via Foresight-Aligned Asynchronous Correction. 
    To address misalignment, we train a lightweight future correction module that predicts future environment representations, allowing the action expert to predict actions from the future time step.
    To reduce reaction time, we introduce confidence-based scheduling optimization that adaptively balances VLM and action expert invocations.
    We also build a llama.cpp-based inference engine tailored for onboard VLA deployment, with system-level optimizations including CUDA graph reuse, GPU-resident intermediate buffering, and flow unrolling.
    Extensive experiments demonstrate that EagleVLA achieves 14.71$\times$ and 9.20$\times$ improvements in control frequency compared with naive PyTorch and vla.cpp on Jetson Orin, while outperforming VLASH by 14.8\% in success rate on LIBERO benchmark.
    The code of our asynchronous algorithm is available on \url{https://github.com/PKU-SEC-Lab/EagleVLA}, and our efficient \texttt{llama.cpp}-based inference engine is available on \url{https://github.com/PKU-SEC-Lab/EagleVLA-Edge}.
\end{abstract}

\keywords{Robotic manipulation,  Edge computing, Asynchronous inference} 

%% file: corl_2026_template_submission/Docs/2intro.tex
\section{Introduction}
Based on Vision-Language Models (VLMs), Vision-Language-Action (VLA) models have demonstrated remarkable performance across a wide range of embodied tasks \cite{black2024pi_0,kim2024openvla}. By leveraging the rich knowledge embedded in pre-trained VLMs and fine-tuning on high-quality robot demonstration datasets, VLA models can directly predict actions chunks from raw observations and natural language instructions \cite{intelligence2025pi_05,kim2025fine}. However, the high parameter count and computational complexity of VLA models pose significant challenges for their deployment on real robotic systems, where low-latency inference is critical yet onboard compute resources are often limited \cite{yue2024deer,yang2026dysl,song2025ceed,su2026execution}.

\begin{figure*}[!tb]
    \centering
    \includegraphics[width=0.75\linewidth]{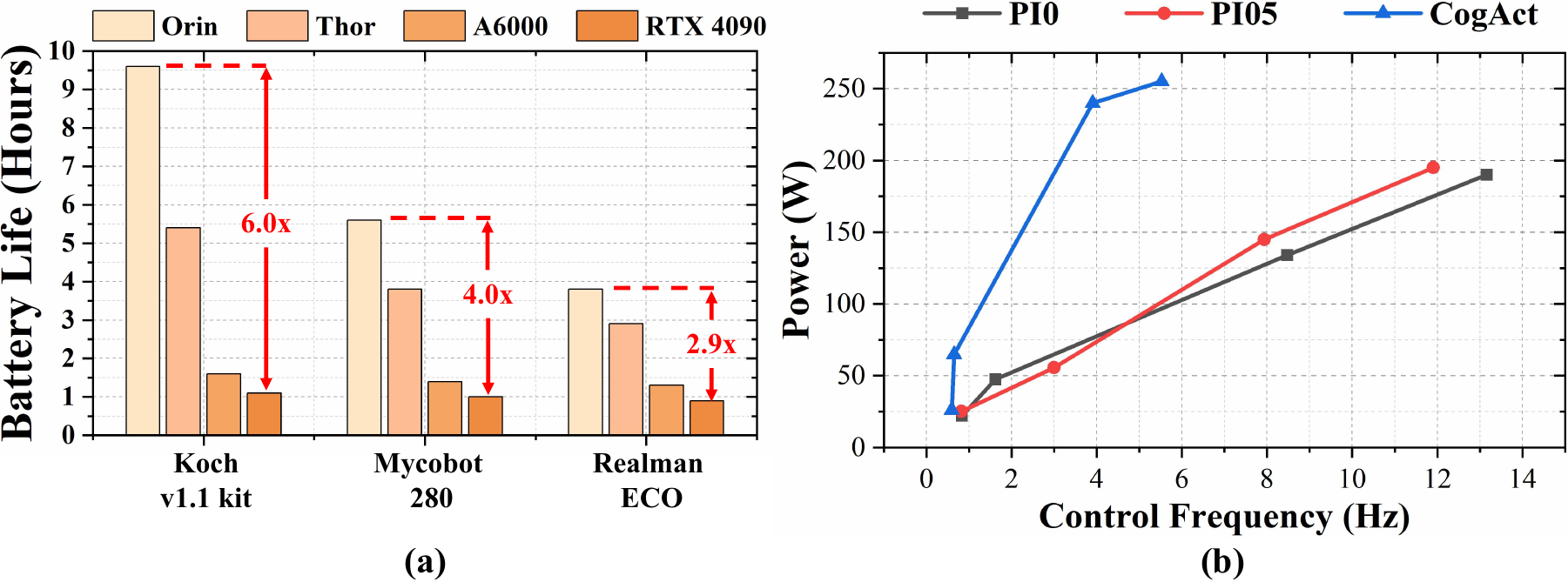}
    \caption{(a) Battery life of three robots equipped with four different computing devices. We use a 500 Wh WILLQ AGV lithium battery pack, and robot's mechanical power consumption is included. (b) Power consumption and control frequency of different VLA inference on four computing devices: Jetson Orin, Jetson Thor, RTX A6000, and RTX 4090.}
    \label{fig:intro:battery_life}
\end{figure*}

To achieve low inference latency and high control frequency, existing works often deploy VLA models on high-end GPUs such as NVIDIA RTX 4090 \cite{black2024pi_0,intelligence2025pi_05,zhang2026foreact,tang2025vlash,niu2026realtime,yang2026realtime}, which provide substantial compute and bandwidth resources. However, these devices introduce severe power consumption issues \cite{yang2024mcubert,zheng2026kerv}. As shown in Figure \ref{fig:intro:battery_life}(a), using an RTX 4090 can reduce battery life by 6.0$\times$ compared to onboard devices like Jetson Orin, significantly limiting robot working time. 
This limits existing VLA works to laboratory demos and makes it difficult to expand its application scenarios \cite{yu2025survey,guan2025efficient,jiang2026fast}.
Even when running such computing devices as online servers, it will also bring extra network latency, and the robot's activity area will be limited \cite{jiang2026fast,yang2026efficientnav}.
Therefore, deploying VLA models on low-power onboard devices is essential for embodied applications \cite{jiang2026fast,hirose2026asyncvla,budzianowski2025edgevla,zheng2026rapid,dai2025actionflow}.

However, due to the limited resources of onboard devices, VLA models suffer from substantial inference latency \cite{jiang2026fast,zhang2026a1,li2025sp}. As shown in Figure \ref{fig:intro:battery_life}(b), state-of-the-art VLA models such as $\pi_{0.5}$ \cite{intelligence2025pi_05} achieve an inference latency of approximately 1.4 seconds on a Jetson Orin, resulting in a control frequency around 0.7 Hz. This low frequency leads to slow reaction to environmental changes and long pauses between action chunks \cite{zhang2026mole,black2026real,black2025training}. Even if asynchronous inference \cite{shukor2025smolvla,zhao2025vla,wei2026f2f} can partially hide inference latency by predicting the next chunk while executing the current chunk, we find that it introduces two critical issues. First, when the VLA finishes predicting an action chunk, the external environment has already changed relative to the image input, leading to a prediction-execution misalignment problem, which becomes even more severe with a long inference time. Second, the robot's reaction speed remains bounded by the VLA inference time, making it difficult to respond in time for complex embodied tasks.

In this paper, we present EagleVLA\footnote{This work was previously referred to as Jetson-PI.}, a method for deploying VLA models on low-power onboard devices via foresight-aligned asynchronous inference. To address prediction-execution misalignment under long inference latency, we train a correction module that predicts future VLM representations conditioned on the actions committed for execution. These predicted representations are then provided as future information to the action expert, enabling it to directly predict actions starting from the future time step. This allows the predicted trajectory to better align with the actual environment at execution time. 
For the slow reaction time issue, we approach the problem from both scheduling and system perspectives. On the scheduling side, as the future correction module can dynamically adjust the foresight horizon, we can invoke the VLM once and subsequently call the action expert multiple times, thus improving control frequency and reducing reaction time. On the systems side, we design an inference framework for VLA on edge platforms based on llama.cpp, and accelerate VLA inference through computation graph reuse and unrolling to cope with the limited bandwidth.
We conduct experiments both in simulation and on real robots across multiple edge devices. On Jetson Orin, EagleVLA achieves 14.71$\times$ and 9.20$\times$ improvements in control frequency compared with naive PyTorch and vla.cpp \cite{nguyen2026vla}. 
And it outperforms VLASH \cite{tang2025vlash} by 14.8\% in average success rate on the LIBERO benchmark \cite{liu2023libero}.


%% file: corl_2026_template_submission/Docs/3background.tex
\section{Background}

\begin{wrapfigure}{r}{0.45\textwidth}
    \centering
    \includegraphics[width=1.0\linewidth]{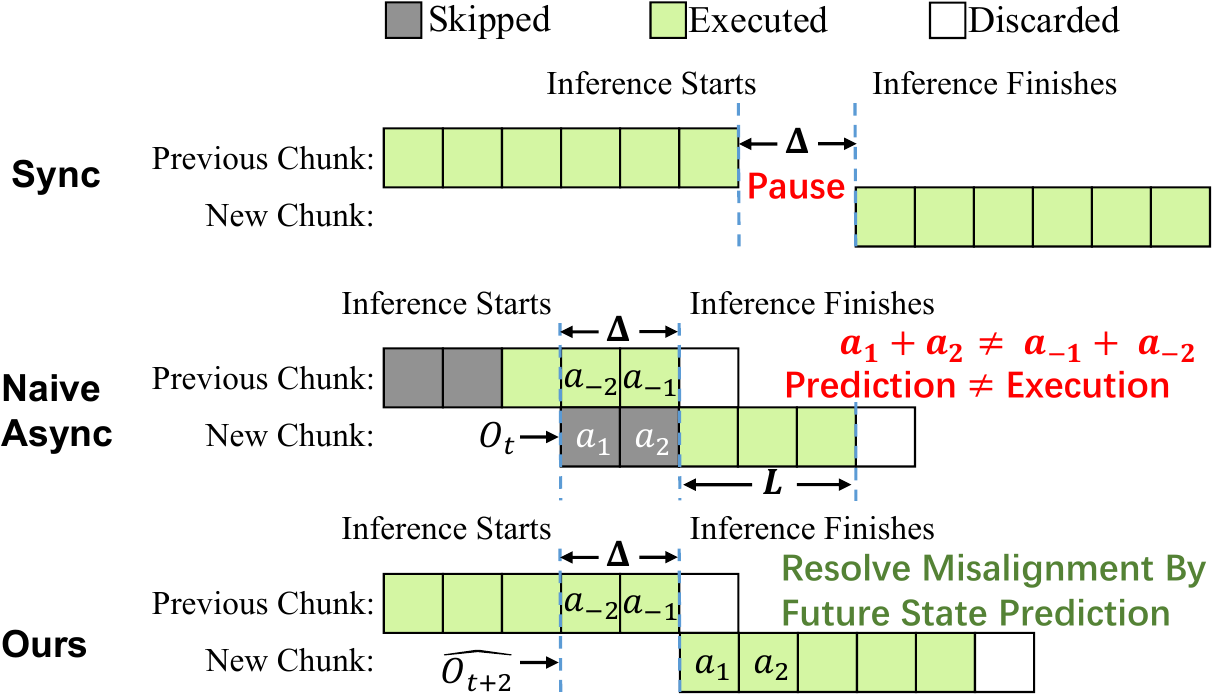}
    \caption{Comparison of different inference modules. $\Delta$ is VLA latency in terms of action steps. $L$ is the number of actions executed in each chunk.}
    \label{fig:back:async}
\end{wrapfigure}

\textbf{Vision-language-action models.}
VLA models have demonstrated strong generalization across a wide range of embodied tasks. State-of-the-art VLA models, such as $\pi_{0}$ and $\pi_{0.5}$, typically adopt a ``VLM + action expert'' architecture \cite{black2024pi_0,intelligence2025pi_05,community2026starvla,bjorck2025gr00t,chen2026dial,liu2026latent,yu2026ac,chen2025fast}. 
During each inference, the current image and language instruction are fed into VLM, which computes and passes the resulting KV cache to action expert. The action expert takes this context along with randomly initialized action noise and uses flow matching to predict an action chunk through multiple denoising steps. The detailed architecture of $\pi_{0}$ and $\pi_{0.5}$ is shown in Appendix \ref{sec:appendix:pi_arch}.
These VLA models are typically deployed under synchronous inference, where the robot must wait for the current action chunk to finish execution before the next inference \cite{black2026real}. Consequently, as shown in Figure \ref{fig:back:async}, the long VLA latency of onboard devices results in long robot pauses between action chunks.

\textbf{Asynchronous VLA Inference.}
As shown in Figure \ref{fig:back:async}, asynchronous inference eliminates pauses between chunks by parallelizing execution of the current chunk with inference of the next chunk \cite{sendai2025leave,zhao2025vla,ye2026world}. 
Among them, SmolVLA implements naive asynchronous inference and directly switches to new action chunks after prediction \cite{shukor2025smolvla}, and RTC mitigates discontinuity between chunks by inpainting the new chunk to produce smoother trajectories \cite{black2026real}. 
However, they still can not solve the prediction-execution misalignment that the environment has already changed relative to the input image of the chunk prediction, which leads to prediction accuracy drop. 
VLASH predicts the robot's state at the end of VLA inference ($q_{t+\Delta}=q_t+a_t+...+a_{t+\Delta-1}$), where $\Delta$ is VLA latency in terms of action steps. 
And uses $q_{t+\Delta}$ to guide the next chunk prediction \cite{tang2025vlash}. But future robot state alone does not reflect environmental changes, and its effectiveness degrades as inference latency increases.
In contrast, we train a future correction module directly predicting future environmental representation. This allows the next chunk prediction to account for how the environment will be affected by the committed actions.

\textbf{World Action Models.}
Another line of work related to ours is World Action Models (WAMs), which leverage world modeling, i.e., predicting future environment, to guide action prediction \cite{yuan2026fast,ye2026world,ye2026gigaworld,xu2026futurevla,zheng2025flare}. 
\cite{intelligence2026pi07,cen2025worldvla} explicitly predict future images, and \cite{yuan2026fast,sun2026vlajepa,su2026wog,zhang2026dreamvla} predict latent representations of future states. 
However, WAMs typically predict environments at the granularity of entire action chunks or sub-tasks, which is usually used as the desired final state and assists in predicting the current action \cite{ye2026world,zhang2026foreact}. 
Our future correction module can be viewed as a lightweight ``world model''. 
The distinction is that our method can dynamically adjust the foresight horizon at action step granularity, which can predict the environment representation after executing the committed actions and assist in predicting the following actions.
This enables addressing the perception-execution misalignment across different computing devices with varying VLA inference latencies.

%% file: corl_2026_template_submission/Docs/4challenge.tex
\section{When Asynchronous Inference Meets Onboard Computation}
We conduct a detailed analysis of asynchronous VLA inference and identify two critical challenges that limit onboard deployment: perception-execution misalignment and long reaction time.

\begin{figure*}[!tb]
    \centering
    \includegraphics[width=1.0\linewidth]{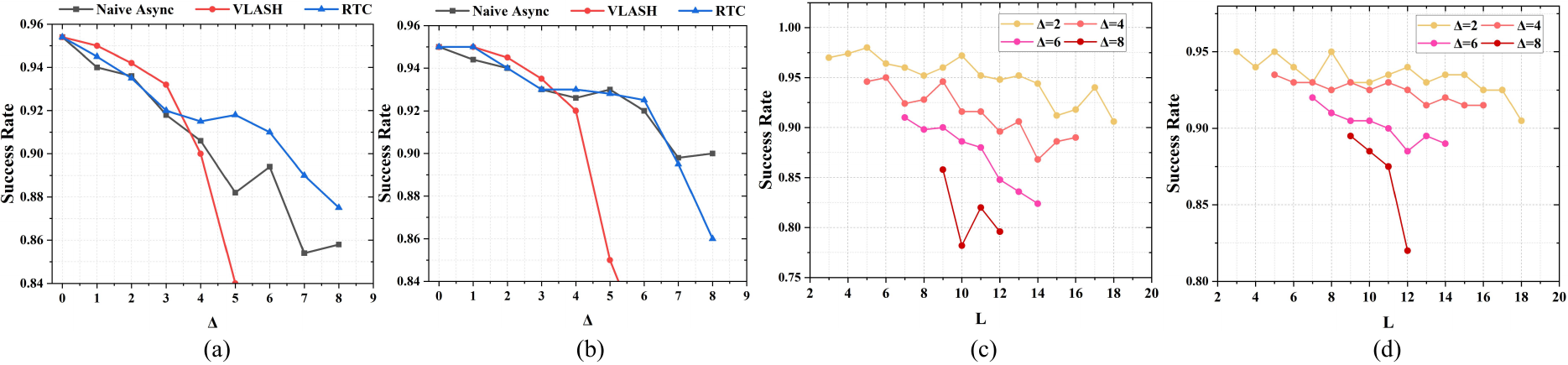}
    \caption{(a)(b) Success rate changes with different $\Delta$ on LIBERO-Spatial and  LIBERO-Goal using $\pi_0$. Here, $L = H - \Delta$, where $H=20$ is the action chunk size. 
    (c)(d) Success rate changes with different $L$ values. using naive asynchronous inference on LIBERO-Spatial and LIBERO-Goal.}
    \label{fig:challenges}
\end{figure*}

\textbf{Perception-execution Misalignment.}
Perception-execution misalignment arises because the environment continues to evolve while the VLA model performs inference for the next action chunk. 
As shown in Figure \ref{fig:challenges}(a)(b), for naive asynchronous and RTC, with VLA inference latency increases, perception-execution misalignment becomes more severe, leading to task success rate drop. 
While VLASH maintains high accuracy with low $\Delta$, for longer latency, predicting future robot state alone is insufficient to capture environmental changes.
Therefore, a method that can provide future environmental information for asynchronous inference on onboard devices is needed.

\textbf{Long Reaction Time.}
Reaction time refers to the interval between an environmental change and the robot's corresponding response. For asynchronous inference, as shown in Figure \ref{fig:overview}, the reaction time in terms of action steps ranges from $\Delta$ to $\Delta+L$ \cite{lu2026faster,shi2026streamingvla}, where 
$L$ is the number of actions executed in each chunk ($L=3$ in Figure \ref{fig:back:async}(b)). As shown in Figure \ref{fig:challenges}(c)(d), increasing $L$ under varying $\Delta$ leads to a drop in task success rate. Because a larger $L$ results in a longer average reaction time. 
In addition, later actions in a predicted chunk tend to be lower quality, as the environment becomes increasingly uncertain when these actions are executed \cite{lu2026faster}. 
To reduce $L$ without introducing pauses between chunks, a straightforward approach is to perform continuous VLA inference (setting $L=\Delta$). However, the reaction time is still bounded by $\Delta$. 
Consequently, the objective of reducing reaction time is transformed into lowering inference latency, especially for onboard deployment.


\textbf{Overview.} 
Figure \ref{fig:overview} presents an overview of EagleVLA. To address prediction-execution misalignment, we propose Foresight-Aligned Asynchronous Correction, which adaptively predicts future environment representation conditioned on the committed actions. This allows the action expert to directly predict actions based on the future environment rather than the outdated observation.
For the long reaction time problem, on scheduling side, we propose Confidence-based Scheduling Optimization, which dynamically adjusts the invocation frequency of the VLM and the action expert based on the confidence of the future environment prediction.
On systems side, we conduct computation graph reuse, intermediate buffering, and unrolling to accelerate VLA inference.

%% file: corl_2026_template_submission/Docs/5method.tex
\section{Method}

\begin{figure}[t]
    \centering
    \begin{minipage}{0.55\textwidth}
        \centering
        \includegraphics[width=\linewidth]{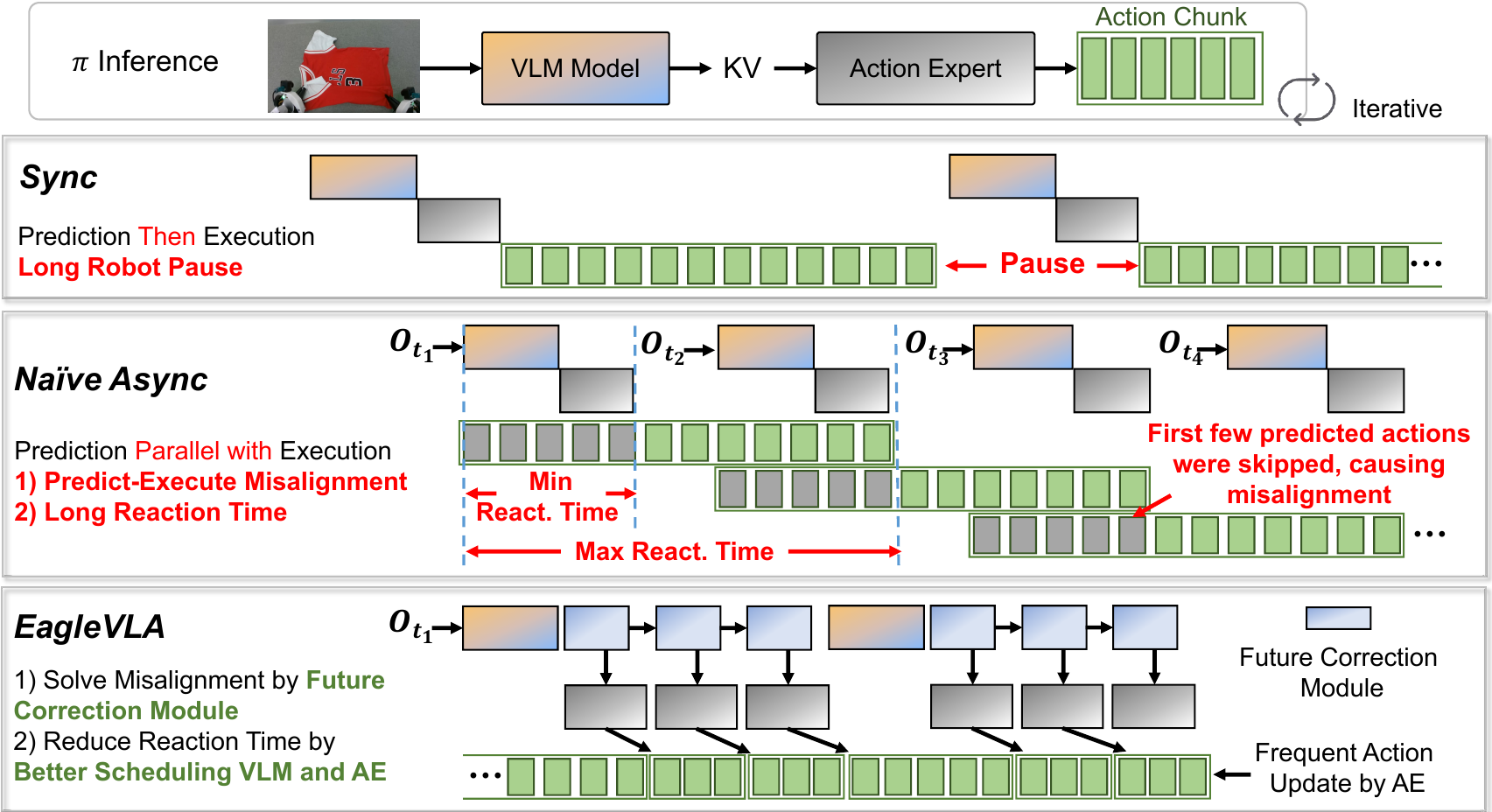}
        \caption{Overview of EagleVLA.}
        \label{fig:overview}
    \end{minipage}\hfill
    \begin{minipage}{0.4\textwidth}
        \centering
        \scalebox{0.65}{
            \renewcommand{\arraystretch}{1.05}
            \begin{tabular}{l|c|cccc}
                \hline
                VLA & Device & ViT  & LLM  & Action Expert & Total  \\
                \hline
                & Orin-30W & 322.4 & 856.4 & 1090.0 & 2268.8 \\
                & Orin-50W & 152.5 & 632.3 & 518.2 & 1403.0 \\
                $\pi_0$ & Thor & 63.7 & 185.0 & 199.2 & 447.9 \\
                & A6000 & 23.4 & 43.5 & 46.9 & 116.1 \\
                & 4090 & 13.1 & 32.2 & 22.4 & 69.5 \\
                \hline
                & Orin-30W & 329.0 & 858.2 & 1102.2 & 2289.4 \\
                & Orin-50W & 152.3 & 631.0 & 536.8 & 1420.8 \\
                $\pi_{0.5}$  & Thor & 61.7 & 187.0 & 209.2 & 457.9 \\
                & A6000 & 23.8 & 51.2 & 50.2 & 128.2 \\
                & 4090 & 13.2 & 35.9 & 25.2 & 76.4 \\
                \hline
            \end{tabular}
        }
        \captionof{table}{Inference latency breakdown (ms) of different VLA models on various devices. We consider different power modes of Jetson Orin.}
        \label{tab:latency_breakdown_vla}
    \end{minipage}
\end{figure}

\subsection{Foresight-Aligned Asynchronous Correction}
\label{sec:method1}




\textbf{Requirements for the future correction module.}
To address the prediction-execution misalignment problem, we train a future correction module that predicts the environment representation at the time when VLA inference completes, guiding next chunk prediction. 
However, the future correction module must satisfy the following requirements: 
(1) Action-conditioned prediction: The module must predict future environment states conditioned on the actions committed to be executed, since these actions directly determine how the environment evolves.
(2) Lightweight design: The module must introduce minimal additional latency, as extra delay would exacerbate the misalignment problem.
(3) Adaptability to varying $\Delta$: as shown in Table \ref{tab:latency_breakdown_vla}, VLA inference latency varies across different hardware platforms and even under different power modes on the same device. The module must adapt to a wide range of $\Delta$ values to remain effective in diverse deployment scenarios.

\begin{figure}[t]
    \centering
    \begin{minipage}{0.55\textwidth}
        \centering
        \includegraphics[width=1.0\linewidth]{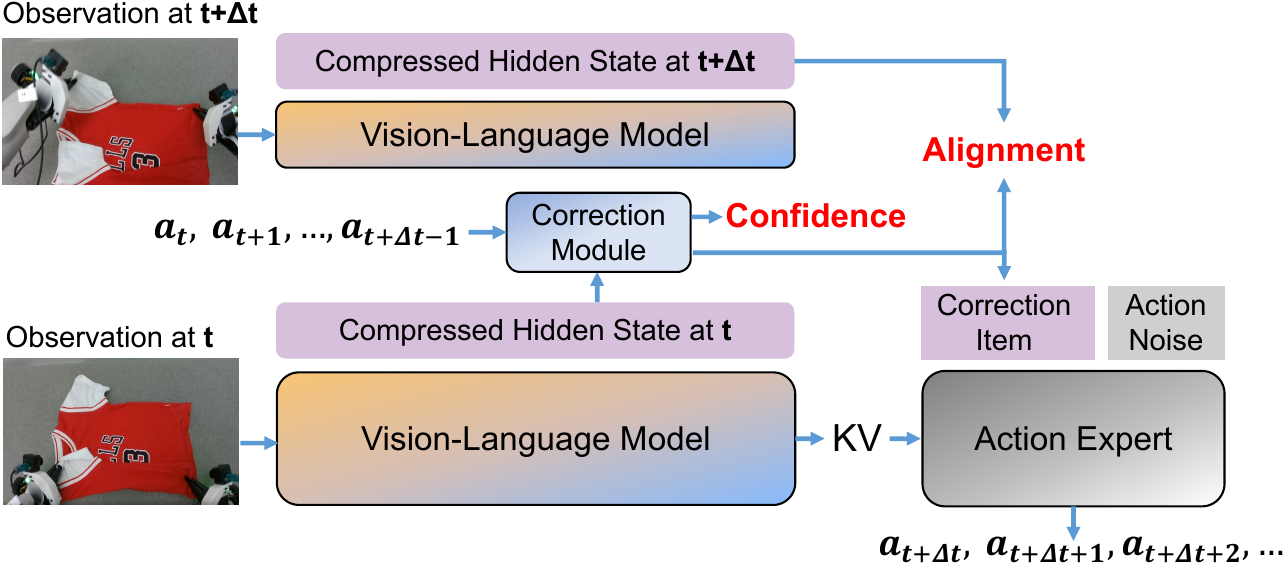}
        \caption{Foresight-Aligned Asynchronous Correction.}
        \label{fig:method1}
    \end{minipage}\hfill
    \begin{minipage}{0.4\textwidth}
        \centering
        \includegraphics[width=1.0\linewidth]{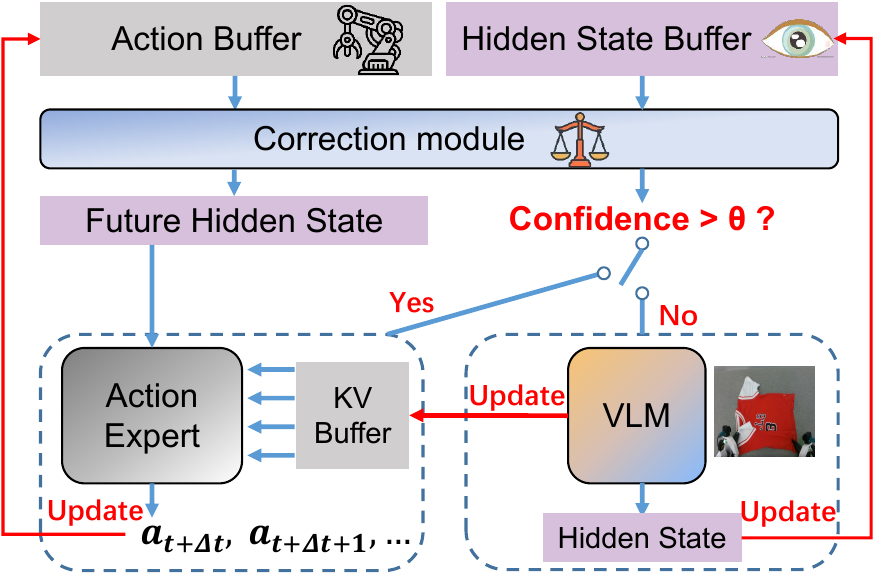}
        \caption{Scheduling Optimization.}
        \label{fig:method2}
    \end{minipage}
\end{figure}

The design of Foresight-Aligned Asynchronous Correction is shown in Figure \ref{fig:method1}. To avoid introducing excessive latency, we do not directly predict future images or correct the KV cache at each layer. Inspired by \cite{su2026wog}, we take the compressed final-layer output of the VLM at $t$ and the committed actions, 
then pass them through future correction module that 
predicts the compressed VLM final-layer output at $t+\Delta$. 
Then pass this lightweight correction item to action expert, enabling action expert to directly predict actions starting from $t+\Delta$. 
The parameter size of future correction module is just 40M, which is only 1\% of the whole VLA and introduces negligible cost.
Architecture details of final-layer state compressor and future correction module are shown in Appendix \ref{sec:appendix:correction_arch}.

To ensure that the future correction module effectively benefits action prediction, the training process must address two objectives: (1) enabling action expert to predict actions starting at $t+\Delta$ with correction item, and (2) enabling future correction module to accurately predict the future environment representation. To this end, we design a two-stage Correction-aware Training pipeline.
In the first stage, we use the ground-truth VLM final-layer state at time $t+\Delta$ and train the compressor and the action expert using the action loss, allowing them to learn how to extract useful information from the future hidden state.
In the second stage, we feed the compressed VLM final-layer output at time $t$ and the action sequence from $t$ to $t+\Delta$ into the future correction module. The module is trained to predict the compressed VLM final-layer output at time $t+\Delta$. Additionally, the module is trained to output a confidence estimate for its future environment prediction, which will later facilitate scheduling optimization. 
The loss function of the second stage is
\begin{equation}
\label{equ:2_stage_loss}
L_{predict} = \| \hat{h}_{t+\Delta} - h_{t+\Delta} \|_2,~~~~~ L_{total} = L_{predict} + \lambda \| \hat{c} + L_{predict} \|_2,
\end{equation}
where $\hat{h}_{t+\Delta}$ and $ h_{t+\Delta}$ is predicted and ground-truth compressed VLM final-layer output at $t+\Delta$ respectively, $\hat{c}$ is the predicted confidence. During training, we randomly sample $\Delta$ to enable the future correction module to adapt to varying inference latencies.

\subsection{Confidence-based Scheduling Optimization}

\textbf{Rethinking the role of VLM and action expert.}
With future correction module, to reduce reaction time, a naive method is to use only the initial image observation, relying solely on the future correction module to predict environmental changes and use the action expert to output all subsequent action chunks without further VLM invocation. 
However, the prediction of future correction module introduces error. These errors accumulate over time and lead to a drop in the success rate \cite{liu2026latent}.
Here we reconsider the roles of the VLM and the action expert after introducing future correction module. The VLM serves to observe current environment and thereby improve the accuracy of the future correction module's prediction. The action expert is invoked to predict actions in real time. Balancing their invocation frequencies is key to achieving low reaction time while maintaining accuracy.


Based on this, we design a Confidence-based Scheduling Optimization approach, where the future correction module serves as the scheduler. Its predicted confidence determines when to invoke the VLM and the action expert. Specifically, as shown in Figure \ref{fig:method2}, when the confidence is above a threshold $\theta$, we consider the predicted future environment information to be sufficiently accurate. So the VLM will be skipped, and the action expert directly predicts actions, improving real-time responsiveness. When the confidence falls below $\theta$, we invoke the VLM to observe the current environment and update both the KV buffer used by the action expert and the hidden state buffer used by the future correction module for subsequent predictions. 
In this way, our method adaptively adjusts the invocation frequency of the VLM and the action expert, reducing reaction time without dropping accuracy. 
We provide an example of confidence evolving over time in experiments and a pseudocode of our algorithm in Appendix \ref{sec:appendix:algorithm}.

\begin{table}[t]
\centering
\caption{Hardware specifications and roofline analysis of different computing devices.}
\label{tab:hardware_specs}
\scalebox{0.65}{
\renewcommand{\arraystretch}{1.2}
\begin{tabular}{l|ccccc}
\hline
 & RTX 4090 & RTX A6000 & Orin 32GB & Orin 64GB & Thor \\
\hline
Max Power (W) & 450 & 300 & 50 & 50 & 130 \\
TOPS & 309 & 309 & 200 & 275 & 510 \\
Memory & 24GB GDDR6X & 48GB GDDR6 & 48GB LPDDR5 & 48GB LPDDR5 & 48GB LPDDR5X \\
Bandwidth (GB/s) & 1024 & 768 & 204.8 & 204.8 & 273 \\
\hline
Balance Computational Intensity & \multirow{2}{*}{618} & \multirow{2}{*}{804} & \multirow{2}{*}{1953} & \multirow{2}{*}{2685} & \multirow{2}{*}{3736} \\
(TOPS/Bandwidth $\times 1000$)&&&&& \\
\hline
VLM & \textcolor{blue}{Compute Bound} & \textcolor{blue}{Compute Bound} & \textcolor{red}{Bandwidth Bound} & \textcolor{red}{Bandwidth Bound} & \textcolor{red}{Bandwidth Bound} \\
Action Expert & \textcolor{red}{Bandwidth Bound} & \textcolor{red}{Bandwidth Bound} & \textcolor{red}{Bandwidth Bound} & \textcolor{red}{Bandwidth Bound} & \textcolor{red}{Bandwidth Bound} \\
\hline
\end{tabular}
}
\end{table}

\textbf{Why don't we parallelize VLM and action expert?} 
\cite{ma2025running} deploys $\pi_0$ on NVIDIA RTX 4090 and parallelizes VLM and action expert inference to reduce reaction time. However, we find this strategy is ineffective for onboard devices. 
We analyze VLA computation on different devices using the roofline model \cite{yuan2026fast,guo2024hg}. The dominant computation in VLA inference is matrix multiplication. For a matrix multiplication operator with input sizes $M\times K$ (activation) and $K\times N$ (weight), the theoretical execution time in FP16 precision is $max(\frac{M*K*N}{compute~throughput},\frac{2*(M*K+K*N)}{bandwidth})$, where $M$ represents token length and $N$ represents model embedding dimension.
Specifically, the inference latency is compute-bound when the first term dominates and bandwidth-bound when the second term dominates. 
The value of $M$ when the two terms are equal corresponds to the balanced arithmetic intensity.
As shown in Table \ref{tab:hardware_specs}, on a high-end GPU such as the RTX 4090, the VLM is compute-bound while the action expert is bandwidth-bound. Therefore, parallelizing the two fully utilizes both compute and bandwidth resources. In contrast, onboard devices typically have much tighter bandwidth. Even VLM computation can not reach the balanced arithmetic intensity, and both the VLM and the action expert become bandwidth-bound. Running them in parallel leads to severe contention for the shared bandwidth, ultimately causing both to slow down significantly.

\subsection{System Design}

\textbf{Unique characteristics of VLA inference.}
Beyond scheduling optimization, we also accelerate VLA inference through system-level design to further reduce reaction time. To the best of our knowledge, no existing deployment framework is specifically designed for edge-side VLA inference. We build our system upon llama.cpp, a popular inference framework for LLMs edge deployment, and adapt it to the unique characteristics of VLA models, which we identify as follows:
(1) \textbf{Deterministic token length.} Unlike LLM decoding, where token length grows progressively, given camera configurations and image resolutions, the token length in VLA inference remains largely constant. The only potential variation is the language instruction, which is typically short and exhibits minimal length fluctuation.
(2) \textbf{Cross-component communication overhead.} A VLA model consists of three components—ViT, LLM, and action expert—that communicate with each other. For example, ViT encodes visual observations and then passes them to LLM, and LLM generates KV caches and passes them to action expert. Traditional inference frameworks write such intermediate results back to CPU memory after GPU computation, introducing unnecessary communication latency.
(3) \textbf{Repeated action expert invocations.} The action expert adopts a flow matching formulation, which requires multiple denoising steps to produce an action chunk. This repeated invocation of the identical computation graph introduces additional latency.

Based on these opportunities, we conduct the following optimizations.
(1) \textbf{Computation graph reuse.} As VLA inference has deterministic token length, we directly reuse the CUDA graph from the first inference to save graph construction time. To ensure that the token length remains constant, we pad the language instruction to a fixed length for every inference.
(2) \textbf{GPU-resident Intermediate Buffer.} Because the sizes of the intermediate results between different components are deterministic, we can directly reserve fixed-size buffers in GPU memory, avoiding writing them back to CPU memory.
(3) \textbf{Flow matching unroll.} We unroll the flow matching computation graph, fusing multiple denoising iterations into a unified computation graph. This reduces the number of graph invocation calls. Although unrolling increases the graph construction time, this overhead is incurred only once during the first inference step, and the graph is reused in subsequent inferences.
Our method is orthogonal to quantization \cite{zhang2026quantvla,zheng2026dyq,park2024quantization,xu2026qvla}, pruning \cite{wang2025specprune,jiang2025better,yang2026efficientvla}, and other acceleration techniques. Here, we focus on system-level optimizations that do not alter the model computation.

%% file: corl_2026_template_submission/Docs/6experiments.tex
\section{Experiments}

We conduct comprehensive experiments in both sim and real to validate the following questions:
\textbf{(1)} How does the performance of EagleVLA compare to other asynchronous inference methods (RTC and VLASH)?
\textbf{(2)} How much improvement in reaction time and control frequency does EagleVLA achieve on onboard devices compared to standard VLA inference?
\textbf{(3)} Does EagleVLA generalize across different inference latencies (varying 
$\Delta$) and different onboard devices (Orin and Thor)?

\begin{table}[t]
\centering
\caption{Success rate comparison across different methods on four sub-datasets of LIBERO, using $\pi_{0.5}$. We report the SR of using Foresight-Aligned Asynchronous Correction alone (ours) and with Confidence-based Scheduling Optimization (+Sched). We estimate inference time of action expert as $\Delta_{ae} = \lceil \frac{\Delta}{3} \rceil$., which is a common ratio both on high-end GPUs such as RTX 4090 and on onboard GPUs such as Orin. We only train one model for all $\Delta$ values.}
\label{tab:main_results}
\scalebox{0.62}{
\renewcommand{\arraystretch}{1.3}
\begin{tabular}{c|cccc|cccc|cccc|cccc}
\hline
 & \multicolumn{4}{c|}{SPATIAL} & \multicolumn{4}{c|}{OBJECT} & \multicolumn{4}{c|}{GOAL} & \multicolumn{4}{c}{LIBERO-10} \\
\cline{2-17}
\hline
Sync. &\multicolumn{4}{c|}{97.3}  & \multicolumn{4}{c|}{99.6}  & \multicolumn{4}{c|}{96.7}  & \multicolumn{4}{c}{93.5}  \\
\hline
$\Delta$ & VLASH & RTC & Ours & +Sched & VLASH & RTC & Ours & +Sched & VLASH & RTC & Ours & +Sched & VLASH & RTC & Ours & +Sched \\
\hline
1 & 98.8 & 97.1 & 97.7 & 98.6 & 99.2 & 98.5 & 98.7 & 98.8 & 96.7 & 96.5 & 97.0 & 97.5 & 94.4 & 92.3 & 93.4 & 93.1 \\
2 & 97.5 & 95.4 & 97.1 & 97.6 & 99.2 & 98.5 & 98.5 & 99.3 & 97.0 & 96.0 & 96.2 & 97.3 & 94.6 & 92.1 & 92.7 & 92.9 \\
3 & 94.4 & 94.5 & 97.1 & 97.2 & 98.8 & 96.4 & 98.2 & 98.4 & 93.3 & 94.3 & 96.7 & 96.4 & 91.9 & 89.6 & 91.3 & 91.9 \\
4 & 92.5 & 92.5 & 96.7 & 96.9 & 96.9 & 97.7 & 97.8 & 98.9 & 93.3 & 93.3 & 95.9 & 96.9 & 89.6 & 85.8 & 91.8 & 92.3 \\
5 & 84.3 & 91.2 & 95.9 & 96.7 & 94.3 & 96.9 & 96.9 & 98.6 & 89.9 & 93.9 & 96.0 & 96.1 & 80.3 & 83.6 & 91.5 & 92.0 \\
6 & 74.4 & 91.7 & 97.0 & 97.5 & 88.5 & 97.4 & 98.3 & 98.4 & 81.3 & 94.0 & 96.7 & 96.9 & 78.5 & 84.6 & 92.2 & 92.9 \\
7 & 51.3 & 91.7 & 97.0 & 97.3 & 81.4 & 97.1 & 98.0 & 98.4 & 76.1 & 92.6 & 97.1 & 97.2 & 77.0 & 85.3 & 92.6 & 92.6 \\
8 & 46.7 & 90.9 & 97.2 & 97.5 & 65.1 & 93.8 & 98.3 & 98.8 & 70.8 & 93.4 & 96.8 & 96.8 & 65.5 & 83.2 & 92.3 & 92.7 \\
9 & 30.1 & 88.8 & 97.0 & 97.3 & 51.7 & 93.1 & 97.5 & 97.6 & 59.7 & 92.5 & 96.3 & 96.5 & 59.6 & 81.0 & 92.0 & 92.2 \\
\hline
Avg & 74.4 & 92.6 & 97.0 & \textbf{97.4} & 86.1 & 96.6 & 98.0 & \textbf{98.6} & 84.2 & 94.1 & 96.5 & \textbf{96.8} & 81.3 & 86.4 & 92.2 & \textbf{92.5} \\
\hline
\end{tabular}
}
\end{table}

\begin{table}[t]
\centering
\caption{Inference latency and control frequency on different devices, evaluated on LIBERO.}
\label{tab:latency_breakdown}
\scalebox{0.67}{
\renewcommand{\arraystretch}{1.2}
\begin{tabular}{lcccccc}
\hline
\multicolumn{7}{c}{\textbf{Jetson Orin}} \\
\hline
Method & ViT (ms) & LLM (ms) & Action Expert (ms) & Total (ms) & Reaction Time (ms) & Control Frequency (Hz) \\
\hline
Naive PI05 & 152.3 & 631.0 & 536.8 & 1420.8 & 1420.8 & 0.70 \\
+Schedule opt. & 152.3 & 631.0 & 536.8 & 1420.8 & 674.9 & 1.48 \\
+Graph reuse  & 71.9 & 196.9 & 143.9 & 412.9 & 184.0 & 5.43 \\
+Intermediate Buffer \& Unroll & 66.6	& 137.0	& 67.6& 271.1 & 97.1 & 10.30 \\
\hline
\multicolumn{7}{c}{\textbf{Jetson Thor}} \\
\hline
Method & ViT (ms) & LLM (ms) & Action Expert (ms) & Total (ms) & Reaction Time (ms) & Control Frequency (Hz) \\
\hline
Naive PI05 & 61.7 & 187.0 & 209.2 & 457.9 & 457.9 & 2.18 \\
+Schedule opt. & 61.7 & 187.0 & 209.2 & 457.9 & 244.7 & 4.08 \\
+Graph reuse & 32.0 & 92.3 & 115.7 & 240.0  & 134.6 & 7.43 \\
+Intermediate Buffer \& Unroll & 29.0 & 54.9 & 48.9 & 132.8 & 61.1 & 16.37 \\
\hline
\end{tabular}
}
\end{table}

\subsection{Simulation Experiments}
\textbf{Experiment Setup.} 
We evaluate EagleVLA on LIBERO \cite{liu2023libero} benchmark based on $\pi_{0.5}$ model. We compare EagleVLA with three baselines: (1) \textbf{Sync.} serves as an optimal baseline and the inferred delay is set to 0 \cite{intelligence2025pi_05}. (2) \textbf{RTC} 
freezes the actions committed to execute and inpaints the rest for smoother trajectories \cite{black2026real}. (3) \textbf{VLASH} estimates future robot state to guide action prediction \cite{tang2025vlash}.

\textbf{Main Results.} 
The simulation results are shown in Table \ref{tab:main_results}. On average over four sub-datasets and different $\Delta$, our method outperforms VLASH and RTC by 14.8\% and 3.9\%. 
Notably, as $\Delta$ increases and the perception-execution misalignment problem becomes more severe, both VLASH and RTC exhibit significant performance degradation.
In contrast, EagleVLA maintains consistently high accuracy, as it adaptively provides future environment information for action prediction across varying $\Delta$.
For $\Delta=9$, EagleVLA shows 45.6\% and 7.0\% on success rate over VLASH and RTC, on average across four datasets, showing the effectiveness of our method.
Compared to synchronous inference, our method eliminates pauses between action chunks with negligible accuracy loss.


\subsection{Latency Evaluation}
Table \ref{tab:latency_breakdown} presents our latency evaluation. Compared to $\pi_{0.5}$ inference on naive PyTorch, our scheduling optimization achieves 2.11$\times$ and 1.87$\times$ reductions in reaction time on Orin and Thor, because action prediction no longer requires invoking the VLM every time. Our graph reuse method achieves a 3.67$\times$ and 1.82 $\times$ reduction in reaction time on Orin and Thor by eliminating per-inference graph construction overhead. Our GPU-resident intermediate buffering and flow matching unrolling achieve 1.89$\times$ and 2.20$\times$ reduction in reaction time on the two devices, by avoiding KV communication and repeated graph invocation across multiple denoising steps. Overall, we achieve 14.71$\times$ and 7.51$\times$ improvements in control frequency on two devices, enhancing the robot's reaction speed and task performance. Compared with vla.cpp, which has a latency of 893.0ms on the Orin platform \cite{nguyen2026vla}, EagleVLA achieves a 9.20$\times$ improvement in control frequency through scheduling and system optimizations.



\begin{figure}[t]
    \centering
    \begin{minipage}{0.5\textwidth}
        \centering
        \includegraphics[width=\linewidth]{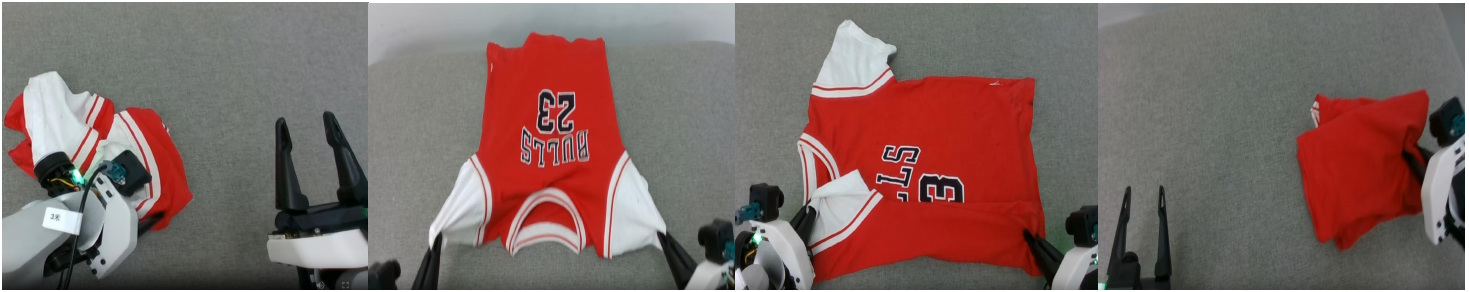}
        \medskip
        \scalebox{0.8}{
            \renewcommand{\arraystretch}{1.2}
            \begin{tabular}{l|c|c|c}
                \hline
                Method & Picking & Folding & Placing \\
                \hline
                RTX 4090 (Baseline)           & 10/10 & 7/10 & 10/10 \\
                Jetson Orin (Naive Async)     & 6/10  & 0/10 & 5/10  \\
                EagleVLA (Ours)              & 10/10 & 8/10 & 9/10  \\
                \hline
            \end{tabular}
        }
        \caption{Real-world results on 3 subtasks (picking, folding, placing) on different deployments.}
        \label{tab:real_world_results}
    \end{minipage}\hfill
    \begin{minipage}{0.45\textwidth}
        \centering
        \includegraphics[width=\linewidth]{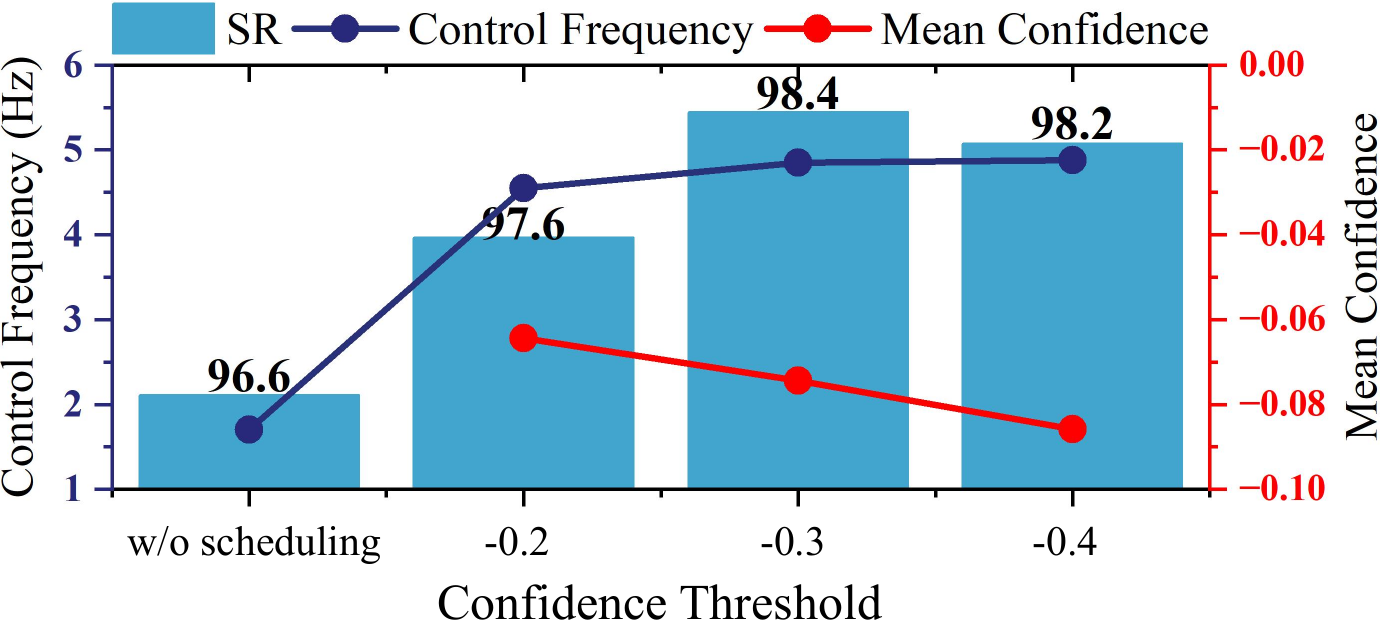}
        \caption{Impact of $\theta$, evaluated on LIBERO-Spatial on Jetson Orin.}
        \label{fig:exp:ablation}
    \end{minipage}
\end{figure}

  \begin{figure*}[!tb]
    \centering
    \includegraphics[width=0.9\linewidth]{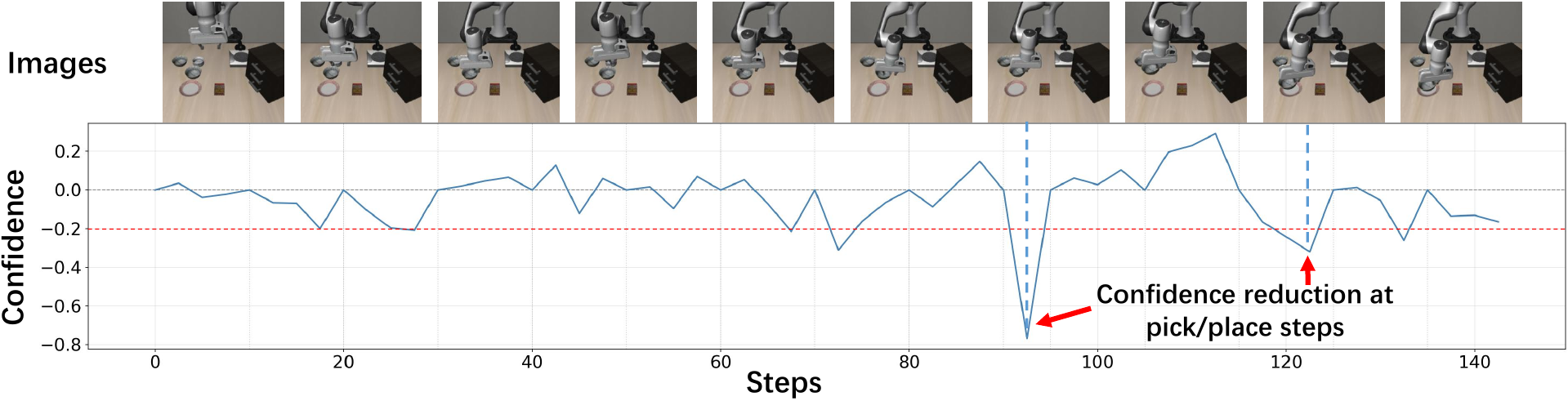}
    \caption{An example of confidence evolution. The confidence threshold is -0.2, and the instruction is ``pick up the black bowl between the plate and the ramekin and place it on the plate''.
}
    \label{fig:exp:confidence_example}
\end{figure*}

\subsection{Real-world Experiments}
Real-world experiments use PrimeBot Research Institute's X2-W robot, equipped with three cameras (one head camera and two wrist cameras, all at 224×224 resolution). We deploy the XR-1 model on a Jetson Orin and have the robot execute a complex task: folding clothes.
This task is divided into 3 subtasks: “put the cloth in the folding area”, “unfold the cloth and fold it neatly”, and “place the folded cloth in storage area”. The robot executes actions at 15 Hz. As shown in Figure \ref{tab:real_world_results}, under constrained compute resources, EagleVLA incurs almost no accuracy loss compared to RTX 4090 deployment, and achieves a clear improvement over naive asynchronous. This shows the capability of EagleVLA to complete complex tasks on edge devices.

\subsection{Analysis of Confidence-Based Scheduling}
\textbf{An example of confidence evolution.}
Figure \ref{fig:exp:confidence_example} presents an example of how confidence evolves during a single task execution. For most of the task duration, confidence remains high, and the future correction module alone is sufficient to predict environmental changes. At critical steps where environment change is hard to predict, such as during grasping and placing, confidence drops and triggers VLM invocation to ensure accurate environment information. This example demonstrates that our method adaptively balances reaction time and environmental accuracy within a single task.



\textbf{Ablation on confidence threshold.}
Figure \ref{fig:exp:ablation} illustrates the impact of the confidence threshold $\theta$ on robot performance. When $\theta$ is large, the VLM is invoked more frequently, leading to a slight decrease in control frequency. When $\theta$ is small, the scheduler invokes the VLM less often, which reduces reaction time but allows slightly larger environmental prediction errors, resulting in a modest drop in average confidence. Nevertheless, across different threshold values, our method achieves improvements in both success rate and control frequency compared to the configuration without scheduling optimization, demonstrating the robustness of our approach.

%% file: corl_2026_template_submission/Docs/7conclusion.tex
\section{Conclusion}
We presented EagleVLA, a method for deploying VLA models on low-power onboard devices. We identified two challenges in asynchronous VLA inference: perception-execution misalignment and long reaction time. To address these, we use Foresight-Aligned Asynchronous Correction that predicts future VLM latents to guide action prediction. We use confidence-based scheduling optimization and system accelerations to reduce reaction time.
Experiments on Jetson Orin show EagleVLA achieves 14.71$\times$ and 9.20$\times$ improvements in control frequency compared with naive PyTorch and vla.cpp with negligible success rate loss.

\section{Limitations}
While EagleVLA successfully enables VLA deployment on low-power onboard devices, significantly improving battery life and activity area compared to high-end GPU setups, the onboard platform remains fundamentally constrained in compute and bandwidth relative to GPU clusters. As model parameter counts and batch sizes continue to scale, the performance gains of EagleVLA may not fully close the gap with high-end GPU deployment. Nevertheless, we believe that for mobile robotics applications where power and portability are critical, EagleVLA offers a practical and efficient solution.

\section{Acknowledgments}
This work was supported in part by New Generation Artificial Intelligence-NationalScience and Technology Major Project under Grant 2025ZD0122502, in part by Beijing Municipal Science and Technology Program under Grant Z241100004224015, in part by NSFC (No. 62495102, 92464104), 
and in part by Shenzhen Key Industry R\&D Project under Grant No. ZDCY20250901105036006, Research and Development of Energy-Efficient Edge Chips for Brain-Cerebellum Coordination in Embodied Intelligence (Project No. 202504012).

%% file: corl_2026_template_submission/Docs/8appendix.tex
\begin{table}[t]
\centering
\caption{Model architecture parameters of $\pi$ series VLA models.}
\label{tab:pi_architecture}
\scalebox{0.8}{
\renewcommand{\arraystretch}{1.2}
\begin{tabular}{lc}
\hline
Parameter & Value \\
\hline
\multicolumn{2}{c}{\textbf{ViT (SigLIP So400m)}} \\
\hline
ViT layers & 27  \\
ViT embedding dimension & 1152 \\
ViT patch size & 14  \\
Input Image Size & 224 \\
\hline
\multicolumn{2}{c}{\textbf{LLM (gemma 2b)}} \\
\hline
LLM layers & 18 \\
LLM embedding dimension & 2048  \\
LLM attention heads & 8\\
LLM head dimension & 256 \\
\hline
\multicolumn{2}{c}{\textbf{Action Expert (gemma 300m)}} \\
\hline
Action expert layers & 18  \\
Action expert embedding dimension & 1024  \\
Action expert attention heads & 8\\
Action expert head dimension & 256 \\
Denoising steps & 10 \\
\hline
\end{tabular}
}
\end{table}

\section{Model Architecture of $\pi$ series Models}
\label{sec:appendix:pi_arch}
Table \ref{tab:pi_architecture} presents the architectural parameters of the $\pi$ series models ($\pi_0$ and $\pi_{0.5}$). The inference pipeline of these models proceeds as follows: 
Given raw observation (2 or 3 images) and a natural language instruction, the ViT encoder first extracts visual features, which are then projected into the token space of the LLM. The LLM processes the concatenated visual and language tokens, providing the KV cache of each layer that serves as conditioning information for the action expert. 
The action expert is a Diffusion Transformer (DiT) whose input is randomly initialized action noise. At each DiT layer, the AE performs cross‑attention where the queries are derived from the action noise tokens, and the keys and values are formed by concatenating the KV caches from the corresponding LLM layer with the action noise tokens.
Through an iterative denoising process based on flow matching, the action expert progressively refines the noise over a fixed number of steps (e.g., 10 steps during inference) to generate a smooth action chunk. This chunk contains a sequence of future actions that can be executed on the robot. Our EagleVLA further introduces a lightweight future correction module that predicts the future VLM hidden states conditioned on the actions committed for execution, enabling the action expert to directly generate actions aligned with the future environment under asynchronous inference.

\section{Architecture of Future Correction Module}
\label{sec:appendix:correction_arch}

\begin{figure*}[!tb]
    \centering
    \includegraphics[width=0.8\linewidth]{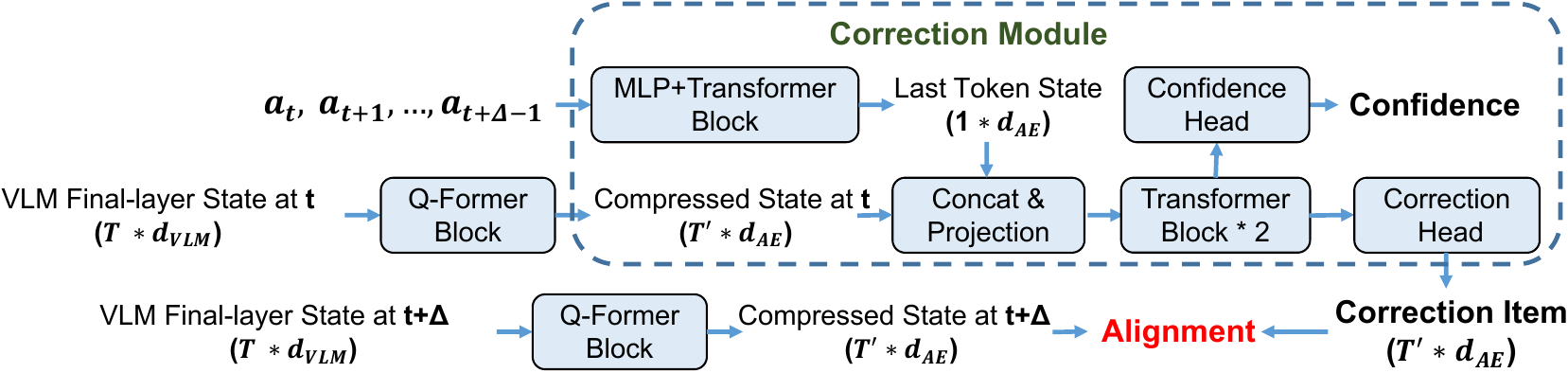}
    \caption{Architecture of future correction module. \(T\) denotes the number of tokens in the VLM hidden sequence; \(T'\) is the compressed token length after Q-Former compression (with \(T' << T\), \(T'= 4\) in practice); \(d_{\text{VLM}}\) is the embedding dimension of the VLM; \(d_{\text{AE}}\) is the embedding dimension of the action expert (DiT); \(t\) and \(t+\Delta\) denotes time steps.  }
    \label{fig:app:correction}
\end{figure*}

As illustrated in Figure \ref{fig:app:correction}, the future correction module takes two inputs:
\begin{itemize}
    \item The sequence of committed actions \(a_t, a_{t+1}, \dots, a_{t+\Delta-1}\) that are guaranteed to be executed during the inference window.
    \item The output hidden state of the VLM final-layer at timestep $t$.
\end{itemize}

The module processes these inputs through a series of components. 
First, the action sequence is passed through an MLP and then into a Transformer block. We take the last token of its output state.
Second, the VLM final-layer output at time \(t\) is compressed by a Q-Former block to produce a compact representation. 
These two branches are then concatenated, projected, and fed into two consecutive Transformer blocks.

From this joint representation, the module splits into two heads:
\begin{itemize}
    \item \textbf{Correction head}: predicts the compressed VLM final-layer hidden state at time \(t+\Delta t\), which is then provided to the action expert as the future environment information. In the training period, the predicted correction item is aligned with the ground-truth compressed VLM final-layer hidden state at time \(t+\Delta t\) as Equation \ref{equ:2_stage_loss}. 
    \item \textbf{Confidence head}: outputs a scalar confidence value \(\hat{c}\) indicating the reliability of the future prediction.
\end{itemize}

The entire module is lightweight, adding minimal computational overhead, and is trained via the two-stage correction-aware training described in Section \ref{sec:method1}.

\section{Algorithm Details of Confidence-based Scheduling Optimization}
\label{sec:appendix:algorithm}
Algorithm \ref{alg:scheduling} shows the details of confidence-based scheduling optimization.

\begin{algorithm}[]
\caption{Confidence-based Scheduling Optimization for Asynchronous VLA Inference}
\label{alg:scheduling}
\begin{algorithmic}[1]
\Require VLM $f_{\text{vlm}}$, action expert $f_{\text{act}}$, future correction module $f_{\text{corr}}$
\Require Initial observation $o_0$, language instruction $l$, confidence threshold $\theta$ ($\theta < 0.0$)
\Require Inference latency of the whole VLA model $\Delta$, inference latency of action expert $\Delta_{ae}$
\Ensure Continuous action execution stream
\State Initialize KV buffer $B_{\text{kv}} \leftarrow \emptyset$, hidden state buffer $B_h \leftarrow \emptyset$, action buffer $B_a \leftarrow \emptyset$
\State Compute initial VLM output: $h_0, KV_0 \leftarrow f_{\text{vlm}}(o_0, l)$
\State Update $B_h$ and $B_{\text{kv}}$: $B_h \leftarrow h_0, B_{\text{kv}} \leftarrow KV_0$
\State Update actions buffer $B_a$: $B_a \leftarrow f_{\text{act}}(h_{0},KV_0)$
\State Initialize parameters $\hat{c} \leftarrow 0.0,~t \leftarrow 0$ \Comment{Initialize a large confidence}
\While{task not finished} \Comment{Actions in $B_a$ are continuously being executed in parallel}
    \If{$\hat{c} > \theta$}
        \Comment{Confidence high: skip VLM, use future correction only}
        \State Retrieve committed $\Delta_{ae}$ actions $\{a_t, a_{t+1}, \dots, a_{t+\Delta_{ae}-1}\}$ from $B_a$
        \State Predict future VLM final-layer state: $h_{t+\Delta_{ae}} \leftarrow f_{\text{corr}}(B_h, a_t, \dots, a_{t+\Delta_{ae}-1})$
        \State Obtain new confidence $\hat{c} \leftarrow f_{\text{corr}}.\text{confidence}()$
        \State Update actions buffer: $B_a \leftarrow f_{\text{act}}(h_{t+\Delta_{ae}},B_{\text{kv}})$
        \State Update hidden state buffer: $B_h \leftarrow h_{t+\Delta_{ae}}$
    \Else
        \Comment{Confidence low: invoke VLM to refresh environment context}
        \State Observe current environment $o_t$
        \State Retrieve committed $\Delta$ actions $\{a_t, a_{t+1}, \dots, a_{t+\Delta-1}\}$ from $B_a$
        \State Update VLM state: $h_t, KV_t \leftarrow f_{\text{vlm}}(o_t, l)$  
        \State Update $B_{\text{kv}}$: $B_{\text{kv}} \leftarrow KV_t$
        \State Predict future VLM final-layer state: $h_{t+\Delta} \leftarrow f_{\text{corr}}(h_t, a_t, \dots, a_{t+\Delta-1})$
        \State Obtain new confidence $\hat{c} \leftarrow f_{\text{corr}}.\text{confidence}()$
        \State Update actions buffer: $B_a \leftarrow f_{\text{act}}(h_{t+\Delta},B_{\text{kv}})$
        \State Update hidden state buffer: $B_h \leftarrow h_{t+\Delta}$
    \EndIf
    \State $t \leftarrow t + \text{(number of executed actions)}$
\EndWhile
\State \textbf{return}
\end{algorithmic}
\end{algorithm}

\begin{table}[t]
\centering
\caption{Training hyperparameters for each stage of our method.}
\label{tab:training_hyperparams}
\renewcommand{\arraystretch}{1.2}
\begin{tabular}{l|c|c}
\hline
Parameter &Stage 1 & Stage 2\\
\hline
Batch size & 32 & 32 \\
Learning rate & 3e-4 & 3e-4 \\
Optimizer & AdamW & AdamW \\
Number of traning steps & 30{,}000 & 25{,}000  \\
Learning rate schedule & Cosine& Cosine\\
Warmup steps & 500 & 500 \\
GPU type & NVIDIA A100 & NVIDIA A100 \\
\hline
\end{tabular}
\end{table}

\section{Training Details}
Table \ref{tab:training_hyperparams} presents our training hyperparameters, including batch size, learning rate, optimizer, number of training steps, and so on.